\documentclass{article}

    \usepackage[preprint]{neurips_2020}

\usepackage[utf8]{inputenc} 
\usepackage[T1]{fontenc}    
\usepackage{hyperref}       
\usepackage{url}            
\usepackage{booktabs}       
\usepackage{amsfonts}       
\usepackage{nicefrac}       
\usepackage{microtype}      
\usepackage{amsmath}
\usepackage{amssymb}
\usepackage{graphicx}

\DeclareMathOperator{\EX}{\mathbb{E}}

\newcommand{\pluseq}{\mathrel{+}=}

\title{Active Measure Reinforcement Learning for Observation Cost Minimization}

\author{%
  Colin Bellinger \\
  Digital Technologies\\
  National Research Council of Canada\\
  Ottawa, Canada \\
  \texttt{colin.bellinger@nrc-cnrc.gc.ca} \\
   \And
   Rory Coles \\
   Department of 
   University of Victoria \\
   Victoria, Canada \\
   \texttt{rfcoles@uvic.ca} \\
   \AND
   Mark Crowley \\
   Faculty of Engineering \\
   University of Waterloo \\
   Waterloo, Canada \\
   \texttt{mark.crowley@uwaterloo.ca} \\
   \And
   Isaac Tamblyn \\
   Security and Disruptive Technologies\\
   National Research Council of Canada \\
   Ottawa, Canada
   Vector Institute for Artificial Intelligence \\
   Toronto, Canada \\
   \texttt{Isaac.Tamblyn@nrc-cnrc.gc.ca} \\
}

\begin{document}

\maketitle

\begin{abstract}

Standard reinforcement learning (RL) algorithms assume that the observation of the next state comes instantaneously and at no cost. In a wide variety of sequential decision making tasks ranging from medical treatment to scientific discovery, however, multiple classes of state observations are possible, each of which has an associated cost. We propose the active measure RL framework (Amrl) as an initial solution to this problem where the agent learns to maximize the costed return, which we define as the discounted sum of rewards minus the sum of observation costs. Our empirical evaluation demonstrates that Amrl-Q agents are able to learn a policy and state estimator in parallel during online training. During training the agent naturally shifts from its reliance on costly measurements of the environment to its state estimator in order to increase its reward. It does this without harm to the learned policy. Our results show that the Amrl-Q agent learns at a rate similar to standard Q-learning and Dyna-Q. Critically, by utilizing an active strategy, Amrl-Q achieves a higher costed return. 
\end{abstract}

\section{Introduction}


When seeing a patient concerned about a potentially cancerous skin blemish, a doctor must decide which diagnostic assessments are required. Some measurements, such as touch and visual inspection, can easily be conducted during the initial consultation, whilst others require sophisticated equipment, drawn-out lab analyses, and have higher costs associated with them. The doctor must actively decide whether the higher cost assessment will provide information necessary in order to accurately and efficiently select the next treatment action.

The above scenario describes a Markov Decision Process (MDP) with observation classes and costs. Environments of this nature include the following properties:
\begin{itemize}
    \item One or more classes of observations (measurements) of the next state are possible;
    \item The measurements have explicit associated costs; and,
    \item The value of the measurement depends on time and space.
\end{itemize} 

Indeed, a wide variety of sequential decision making tasks, such as materials design, public health planning during a pandemic and operational planning (business decision making) involve the choice of actions and classes of observations with associated costs. 

The MDP formalism and the environments on which reinforcement learning (RL) algorithms are developed and tested, however, are not designed to explore such settings \cite{brockman2016openai}. In the canonical framework, observations of the state of the environment are produced automatically, instantaneously and have no explicit associated costs. Generally, agents are agnostic to the state observations provided by the environment in the sense that they learn from what they receive. To the extent that the agent might try to improve the quality of observations, it is through deep feature representations \cite{mnih2015human}, maintaining a belief state for partially observable MDPs \cite{kaelbling1998planning}, or taking actions to change the state of the environment in order to gain a better understanding of it \cite{bossens2019learning}. Thus prior work has considered observations, yet has not dealt with the selection of observation classes, nor minimizing observation costs.

Here, we frame MDPs with observation classes and costs as an active learning problem. Active learning is typically applied to supervised machine learning with the aim of reducing the cost of labelling training data \cite{Settles2012}. However, active learning has recently been applied to RL in the context of determining reward from external experts \cite{epshteyn2008active,krueger2016active,schulze2018active}. Conversely, we postulate that in some domains observations of the state of the environment, like supervised labels, are expensive to obtain. In the context of this work, the active component is applied to learning which measurements to make in a given state at a particular time, or deciding not to make a measurement at all - thereby foregoing the additional information and cost associated with it. The aim is to discounted sum of rewards minus observation costs, which we denote as the \emph{costed return}. 

We propose the Active Measure Reinforcement Learning (Amrl) framework in which the agent learns a policy and a state estimator in parallel via online experience. The agent chooses actions pairs that change the environment and dictate whether the next state is measured directly or estimated. As the state estimator is refined over time, the agent smoothly shifts to increasingly rely on it thereby lowering its observations cost. This enables the Amrl agents to achieve a higher costed return.

We demonstrate an implementation of Amrl using Q-learning and a statistical state estimator (Amrl-Q). We compare Amrl-Q to Q-learning and Dyna-Q on four benchmark learning environments, including a new chemistry motivated environment; specifically, the junior scientist environment. The results show that Amrl-Q achieves a higher sum of rewards minus observation cost than Q-learning and Dyna-Q, whilst learning at an equivalent rate to Q-learning and Dyna-Q. 


\subsection{Contributions}

The main contributions of this work are:
\begin{itemize}
    \item Formalization of MDPs with observation classes and costs
    \item Definition of the Active Measure RL framework (Amrl)
    \item Initial implementation of a Q-learning approach, the Amrl-Q algorithm
    \item Analysis of Amrl-Q on benchmark RL environments
\end{itemize}

\section{Related Work}

Previous work on active reinforcement learning has focused on ameliorating the problem of defining a complete reward function over the state-action space \cite{akrour2012april,krueger2016active,schulze2018active}. In addition to selecting an action at each time step, the agents in these proposals actively decide to request a human expert to provide the reward for the state-action pair. To minimize reliance on human experts, there is a cost assigned to requesting a human-specified reward. The agent aims to minimize this cost whilst maximizing the discounted sum of rewards. Similarly, Amrl maximizes the discounted sum of rewards minus the sum of observations cost. However, the Amrl agent differs in the sense that state observations are the bottleneck in the learning process rather than the rewards. Moreover, the Amrl agent may have multiple different measurements of the state of the environment available to it, each of which has a distinct cost. 

Active perception relates to our work in that the agent takes actions to increase the information available \cite{gibson1966senses}. The key distinction, however, is that in active adaptive perception applied to RL, the agents employ self-modification and self-evaluation to improve its perception \cite{bossens2019learning}. Alternatively, the Amrl agent aims to judiciously select observation classes in order to have the necessary and sufficient amount of information to choose the next action in order to maximize costed return.   

Recently, the authors in \cite{chen2017double,li2019multi} proposed the extension of the concept of multi-view learning from supervised domains reinforcement learning. They formulate this as an agent having multiple views of the state-space available to it. This is the case, for example, for agents controlling autonomous vehicles equipped with multiple sensors. This previous work, however, does not contain the concept of observation costs, which are fundamental in applications of Amrl. 
Approximate dynamic programming (ADP) aims to ameliorate the ``curse-of-dimensionality'' in dynamic programming \cite{powell2007approximate}. It is connected to our work in the sense that it introduces a new component, the post-decision state, to the interaction with the environment. Alternatively, our work, which is not focused on the curse-of-dimensionality, formulates an action pair that determines the process to be applied in the environment and the class observation to be made.

The learning of the state transition dynamics of the Amrl framework is consistent with the techniques employed in model-based RL \cite{deisenroth2011pilco,kumar2016optimal,gal2016improving}. The goal of model-based RL, however, is to reduce the number of real-world training steps needed to obtain an optimal policy. This does not solve our problem of selecting the observation class, nor minimizing associated observations. 




Learning algorithms for POMDPs utilize a state estimator to internalize the agent's recent experience in order to reduce uncertainty in partially observable environments. At each time step, the next action $a_t$ is selected based on the the agent's belief state $b_t$ as determined by its state estimator, rather than the observation emitted from the environment $o_t$ \cite{kaelbling1998planning}. Alternatively, in Amrl the agent is learning an optimal policy under a MDP with observation costs. The agent chooses between paying the cost to measure the true state of the environment $s_t$ or estimating it $\hat{s}_t$. Thus, in Amrl the state estimator is a mechanism to increase the costed return, not manage partial observability. 


\section{Preliminaries}

We define active measure reinforcement learning as a tuple: $(S, A, P, S^\prime, R, C, \gamma)$. The components $(S, A, P, S^\prime, R, \gamma)$ make up a standard MDP where $S$ is the state-space, $A$ is the action-space, $P(s^\prime | s, a)$ is the state transition probabilities, $R(s, a)$ is the reward function, and $\gamma \in [0,1]$ is a discount factor. $P$ and $R$ are not known by the agent. $C(m)$ is the cost charged to the agent each time it decides to measure the state of the environment. Thus, for a state $s$, the environment returns the cost as follows:
\begin{equation}
    C(m)= 
    \begin{cases}
        c>0,& \text{if } m = \text{measure the state}\\
        0,              & \text{otherwise}.
    \end{cases} 
\end{equation}

Applications may have multiple observation classes $\mathcal{M} = \{0, 1,2,...\}$, such as different sensors that serve different purposes. In this case, each measurement class $m\in \mathcal{M}$ may have a different associated cost. Selecting $m$ constitutes the active learning choice on the part of the agent. The values of $m>0$ indicate a specific observation class (such as a specific sensor) to be used, whereas $m=0$ specifies that no measurement of the environment is to made\footnote{When $m=0$, the agent uses its state estimator in place of a measurement of the environment.}. 



As in \cite{schulze2018active}, at each time step $t$ the agent selects an action pair. In Amrl, the action pair $(a_t, m_t)$ consists of an atomic process $a_t \in A$ (\textit{e.g.}, move left) and an observation class $m_t \in \mathcal{M}$. Thus, if $m_t>0$, the process $a_t$ is applied to the environment, and the environment returns the reward $r_{t+1}$ and the next state observation $s_{t+1}$ measured via $m_t$ ($r_{t+1}, s_{t+1} = Env(a_t, m_t)$). Here, $s_{t+1}$ results from the underlying, unknown transition dynamics $P(s_t,a_t)$. For $m_t=0$, the process $a_t$ is applied to the environment, but the environment only returns the reward $r_{t+1} = Env(a_t, m_t)$. In this case, the Amrl agent estimates the next state $\hat{s}_{t+1} \sim \hat{P}(s_t,a_t)$, and selects its next action pair $(a_{t+1}, m_{t+1})$ based on this estimate, $\hat{s}_{t+1}$. This leads to an alternative agent-environment interaction sequence of the form:
\begin{equation}
    s_0,a_0,(r_1-c_1),\hat{s}_2,a_1,(r_2-c_2),\hat{s}_3,a_3,(r_3-c_r),s_4...,
\end{equation}
where the agent starts each episode with a true measurement of the environment's current state, $s_0$, and proceeds to sequentially select action pairs that determine the process $a_t$ to be applied and whether to measure the next state $s_{t+1}$ or estimate $\hat{s}_{t+1}$ instead. 

Importantly, the reward emitted from the environment is always a function of the process $a_{t}$ and the true state of the environment $s_t$ irregardless of whether the agent selected $a_t$ based on $s_t$ or an estimate $\hat{s}_t$. For simplicity and generalization, at times we drop the hat notation on the state estimates.

In this work, we focus on episodic environments with discrete states, $S = \{1,...,|S|\}$, and action sets $A = \{1,...,|A|\}$, and stationary state-transition dynamics. In an MDP with measurement costs, the objective is to select a sequence of action pairs $(a_t,m_t)$ that maximize the costed return, which is defined as the discounted sum of rewards minus the sum of measurement costs:
\begin{equation}
    v(s) = \EX \bigg[ \sum^{\infty}_{t=0} \gamma^t \big(R(s_t, a_t) - C(m_t)\big) ~|~ s=s_0\bigg].
\end{equation}

In Amrl, a policy, $\pi$ maps states $S$ and actions pairs $AP$ to a probability $\pi: S \times AP \rightarrow [0,1]$, such that $\pi(s,ap)$ is the probability of selecting action pair $ap\in AP$ in state $s\in S$. The value function associated with policy $\pi$ is:
\begin{equation}
    v_\pi(s) = \EX \bigg[ \sum^{\infty}_{t=0} \gamma^t \big(R(s_t, a_t) - C(s_t, a_t)\big) ~|~ s=s_0\bigg],
\end{equation}
where the actions are selected according to $\pi$. Since actions pairs can be though of as a higher-level class of action, the standard RL theorems hold. Thus, there is at least one policy $\pi^*$ such that $V^{\pi}(s) \ge V^{\pi^*}(s)$, where $\pi^*$ is an optimal policy and $V^*$ is the corresponding value function.

\section{Amrl-Q} \label{sec:Amrlq}

We propose an initial implementation of the Amrl framework for a tabular learning environment. Our proposed solution utilizes Q-learning for the value function and a statistical state transition model. We focus on tabular problems here for clarity in the demonstration and analysis. Our future work will implement Amrl solutions for continuous state and action spaces.

\subsection{Overview}

\begin{figure*}
\centering
    \includegraphics[scale=0.25]{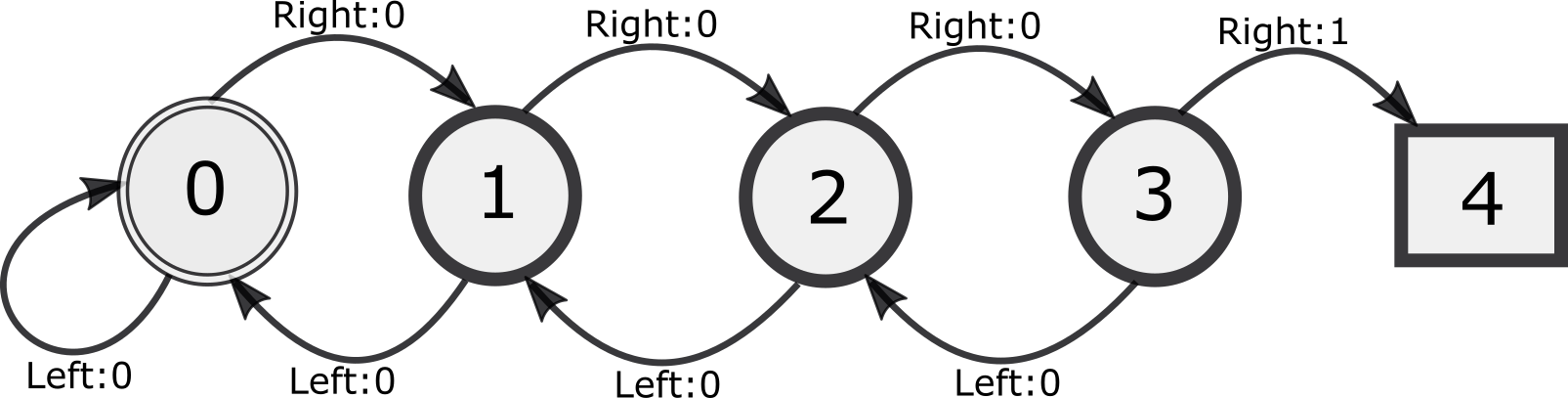}
\caption{This figure illustrates a five state Markov chain.}
\label{fig:markovChain}
\end{figure*}

As previously stated, Amrl-Q framework learns a value function $Q$, and a state estimator $\hat{P}(S_{t+1}|S_t,a_t)$ in parallel. Learning $\hat{P}$ and $Q$ is essential to the active learning based solution which enables the agent to reduce its the total number of times it requests a true measurement. The theory behind this can be demonstrated with the Markov chain in Figure \ref{fig:markovChain}.

This Markov chain forms a two action (left, right) episodic RL problem where the agent starts in stage zero, and receives a reward of one upon entering the absorbing state, state four. For temporal difference (TD) learning methods, such as Q-learning, applied to episodic problems such as this, the value of states and actions is refined over episodes of training from the state closest to the absorbing state back to the start state. The backup algorithm for Q-learning is:
\begin{equation}
    Q(s_t,a_t) \leftarrow Q(s_t,a_t) + \alpha \bigg[r_{t+1} ~ \gamma \max_a Q(s_{t+1},a) - Q(s_t,a_t) \bigg],
\end{equation}
where $Q(s_t,a_t)$ is the value of action $a$ in state $s$ at time $t$, $\alpha$ is the learning rate and $\gamma$ is the discount factor. If we assume a $Q$-table initialized to zeros, after one episode of training is complete, only $Q(s=3,a=\text{right})$ will have a value greater than zero; after the second episode is complete, states 2 and 3 will have values greater than zero, and so on. In general, for an $n$-state chain of this nature, the agent will require $n-1$ episodes of training to start to improve the $Q$-values associated with the start state, state 0. 

The number of times the agent visits each state per episode indicates how many true measurements of the environment it will make. We can estimate this by calculating the fundamental matrix $N$ of the absorbing Markov chain $P$ shown in Figure \ref{fig:markovChain}. The fundamental matrix is defined as $N=(I-Q)^{-1}$, where $I$ is the identify matrix and $Q$ is the $t\times t$ matrix representing the transient states in $P$. Based on this, the expected number of state visits before absorbing for an agent starting in state 0 and following a random policy is 8,6,4 and 2, respectively. Thus, in the first four episodes of training, the Q-agent is expected to take 46 measurements of the environment.   

If we consider the state estimator $\hat{P}$ learned by Amrl, according to the calculations above, in the first episode of training the agent is expected to have tried both actions in each state 4, 3, 2 and 1 times, respectively. Since for a deterministic $P$, the agent must try each state-action pair once to have an accurate $\hat{P}$, Amrl can safely switch from actively measuring the next state, to estimating it with $\hat{P}$ after the first episode of training. Moreover, because the agent tries actions in states closer to the start sooner and more frequently, it can switch to using $\hat{P}$ in these states even before the first episode of training ends. In this way, Amrl is able to improve measurement efficiency well beyond what can be achieved by standard RL methods and model-based RL. 

\subsection{Algorithm}

The Amrl-Q algorithm maintains $|A|$, $|S|\times|S|$ count-based statistics table for state transitions models $\hat{P}_a$. In this initial presentation, we limit the agent to selecting from one observation class. Therefore, the agent maintains an $|S|\times(|\mathcal{A}| \cdot 2)$ dimensional $Q$-table, where $|\mathcal{A}| \times 2$ is the number of action pairs. An environment with 2 action has 4 action pairs, and thus, a four column $Q$-table.

The Q-table is update in the standard way as:
\begin{equation}
    Q(s_t,a_t) \leftarrow Q(s_t,a_t) + \alpha \bigg[(r_{t+1}-c_{t+1}) ~ \gamma \max_a Q(s_{t+1},a) - Q(s_t,a_t) \bigg]
\end{equation}
The agent employs an $\epsilon$-greedy strategy to pick action pairs from the Q-table. If the action pair at time $t$ includes $m_t = 1$, then the agent chooses to pay the cost $c$ of measuring the next state from the environment. Otherwise, the agent estimates the next state from its model as $s_{t+1} \sim \hat{P}_a(s_t)$. When the agent chooses to measure the true state, it updates $\hat{P}_a(s_t,s_{t+1})$ for the corresponding action $a=a_t$

Much like a human learning a new task, the first few times an agent enters a state it must measure the result of taking an action. We produce this behaviour by initializing Q-values for action pairs involving state measurements $m=1$ with small positive value, and zero for Q-values related to measurements $m=0$ (implications of initialization are discussed below).

Over successive visits to a state $s$ and applications of a process $a$ and measurement $m=1$, the return for $s$ will be less than the maximum possible return because the measurement cost $C(m)$ is subtracted from the reward $R(s,a)$. Since moving without measuring does not incur an additional measurement cost, in time and as the model improves, moving and relying on the learned model produces an increased reward and the agent shifts to this strategy.

The outline of the algorithm is:
    \begin{itemize}
        \item Initialize a biased Q-table of size $|S|\times |A|\cdot 2$
        \item Initialize |A| state-transition statistic table of size $|S|\times |S|$ to zeros
        \item get the first state $s = s_0$ from the environment
        \item repeat until done
        \begin{itemize}
            \item Select action pair $(a, m)$ with $\epsilon$ greedy policy from Q table for state $s$
            \item Apply action $a$ to environment
            \item If measure $m=1$:
            \begin{itemize}
                \item Measure next state $s^\prime$ in environment
                \item Update state transition model for action $a$ $\hat{P}_{a}[s, s^\prime] \pluseq 1$ 
            \end{itemize}
            \item Else:
            \begin{itemize}
                \item Sample next state $s^\prime \sim \hat{P}_{a}(s)$
            \end{itemize}
            \item Get reward $r$ from environment
            \item Get cost $c$ from the environment
            \item Update Q table for state $s$ with tuple $(s, a, r-c, s^\prime)$
            \item Set $s \leftarrow s^\prime$
        \end{itemize}
    \end{itemize}

\section{Experimental Setup}

The following experiments are conducted on episodic, discrete state and action problems. Our analysis involves three standard RL environments (Chain, Frozen Lake $8\times 8$ and Taxi) and one new environment (Junior Scientist). Each of these environments has the feature that the agent must actively decide if and when to measure the state of the environment. In the case of the OpenAI Gym environments (Frozen Lake and Taxi), we have implemented a wrapper class in Python that adds the Amrl functionality. 

\subsection{RL Environments}

\textbf{\emph{Chain environment}}: A chain of 11 states, $s \in \{0,...,11\}$, where the agent starts at $s_0$ and the episodes ends when the agent enters $s_{10}$. Upon entering goal state $s_{10}$, the agent receives a reward of $r=1$. The agent receives a reward of $r=-0.01$ at each time step. The agent is charged a measure cost of $c=0.05$ for measuring the state of the environment. Measuring the state results in the environment returning the current state in chain. The action space is $A=\{\text{move left, move right}\}$. We evaluate the performance with both deterministic state transitions and stochastic state transitions. In the stochastic setup, the environment has a probability $p$ of the actions being swapped at each time step.

\textbf{\emph{Frozen Lake $8 \times 8$ environment}}: In this environment, the agent learns to navigate from a start location to a goal in a frozen lake grid with holes in the ice. Each episode ends when the agent reaches the goal or falls through a hole in the ice. The agent receives a reward of $r=1$ at the goal, $r=0$ otherwise. The agent pays a cost of $c=0.01$ for measuring the state of the environment. Measuring the state results in the environment returning the current position in the 2-dimensional frozen lake grid. The action-space in the environment is $A=\{\text{move left, move right, move up, move down}\}$. In this implementation, the agent is prevented from moving off the grid. We evaluate the agents with both the predefined deterministic and slippery settings in the openAI gym.

\textbf{\emph{Taxi environment}}: The agent learns to navigate a city grid world to pick up and drop off passengers at the appropriate location \cite{dietterich2000hierarchical}. The agent receives a reward $r=20$ for dropping off at the correct location, $r=-10$ for illegal pickup or drop-off and $r=-1$ at each time step. The agent is charged a cost of $c=0.01$ for measuring the state of the environment. Measuring the state results in the environment returning the current position in the city grid. The action-space includes $A=\{\text{move left, move right, move up, move down, pickup, drop-off}\}$.

\textbf{\emph{Junior Scientist environment}}: This environment emulates a student learning to manipulate an energy source to produce a desired state change in a target material. Specifically, the agent starts with a sealed container of water composed of an initial $h_0$ percent ice, $l_0$ percent water and $g_0$ percent gas ($h_0+l_0+g_0=1$). The agent learns to sequentially and incrementally adjust a heat source in order to transition the ratio of ice, liquid, gas from $(h_0,l_0,g_0)$ to a goal ratio $(h,l,g)$. The episode ends when the agent declares that it has reached the goal and it is correctly in the goal state. The action-space includes $A=\{\text{decrease, increase, done}\}$, where \emph{decrease} and \emph{increase} are fixed incremental adjustments in the energy source. The agent receives a reward of $r=1$ when it reaches the goal and it correctly declares that it is done, and receives a reward of $r=-0.05$ at each time step. The agent is charged $c=0.01$ for measuring the state of the environment. Measuring the state results in the environment returning the cumulative energy which has been added or removed from the system.

\subsection{RL Algorithms}

We compare the relative performance of Amrl-Q to non-active methods: Q-learning \cite{watkins1992} and Dyna-Q \cite{sutton1990}. Since neither Q-learning nor Dyna-Q are active RL methods, they require a measurement of the environment at each time step. As a result, they are charged the measurement cost $C$ at each time step. The relative performance of these methods is assessed in terms of the sum of reward minus observation costs, along with the mean number of steps and measurements per episode. To the best of our knowledge, we are proposing the first solution to the Amrl problems. As such, Q-learning and Dyna-Q are a reasonable baseline for comparison in this introductory work.

\subsection{Methodology}

For each RL algorithm in our evaluation, we utilize a discount factor of $\gamma=0.9$ and $\epsilon$-greedy exploration $\epsilon=0.1$. The $Q$-tables for both Q-learning and Dyna-Q are initialized to zeros. The columns of the $Q$-table in Amrl-Q associated with estimate ($m=0$) are initialized to zeros and those associated with measure ($m=1$) were set to a small positive, typically $0.1$ (we also explore the impact of larger values). The results presented are mean performance averaged over 20 random trials, enough to be statistically significant. We employ 5 planning steps, a reasonable baseline, in Dyna-Q after each real step. 

\section{Results}

We initially focus on the performance of each agent in the deterministic environments. We highlight the impact of stochasticity in the Discussion.

\subsection{Chain}
\begin{figure*}
\centering
    \includegraphics[scale=0.4]{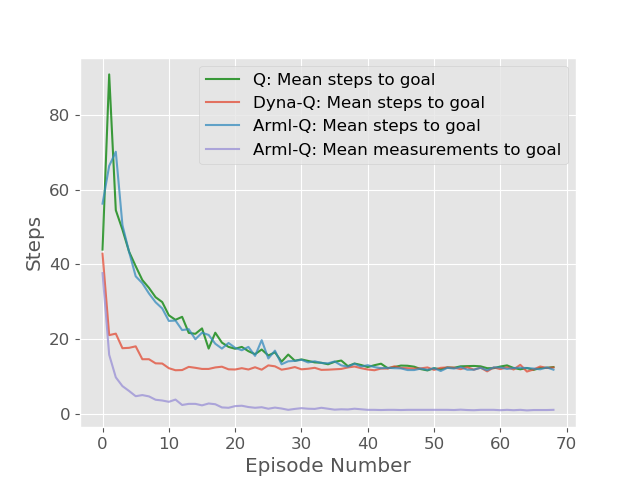}
    \includegraphics[scale=0.4]{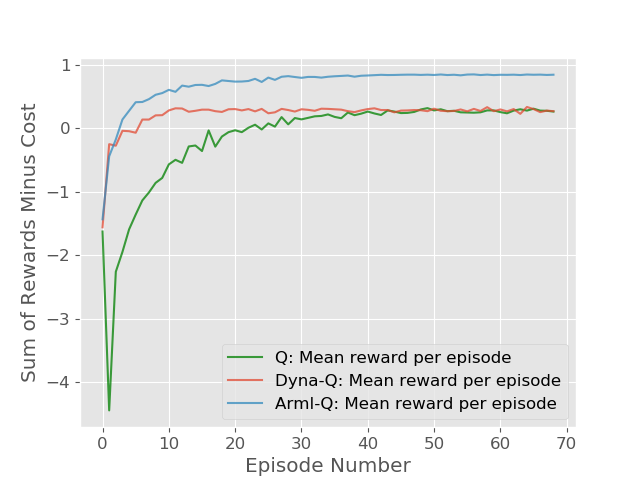}
\caption{Left: Mean steps to the goal by episode in the deterministic Chain environment. Right: Mean costed return in the deterministic Chain environment.}
\label{fig:detChainStepsAndSum}
\end{figure*}

The mean performance of each agent is shown in Figure \ref{fig:detChainStepsAndSum}. The left plot displays the mean number of steps to the goal for Q-learning, Dyna-Q, and Amrl-Q. All three methods learn a policy that takes a similar number of steps to the goal. Naturally, Dyna-Q learns faster (red line versus green and blue). It worth noting that Dyna styled planning could easily be incorporated into Amrl-Q as an enhancement, however, this is beyond the scope of this study.

Whilst Q-learning and Dyna-Q require a measurement after each action, Amrl-Q actively decides whether to measure or estimate the next state. The purple line in Figure \ref{fig:detChainStepsAndSum} show the mean number of measurements per episode made by Amrl-Q. In the very early episodes, the number of measurements is similar to Dyna-Q, however, it quickly drops well below the alternatives.

The cost savings resulting from fewer measurements for Amrl-Q can be seen in the higher costed return presented in the plot on the right. As in the previous analysis, Amrl-Q is initially similar to Dyna-Q. This holds while Amrl-Q learns about state transition dynamics. Because Amrl-Q dynamically shifts its measurement behaviour in each state as it learns about the transition dynamics, over episodes of training it reduces its measurement costs to acquire a higher costed return (blue line).

\begin{figure*}
\centering
    \includegraphics[scale=0.25]{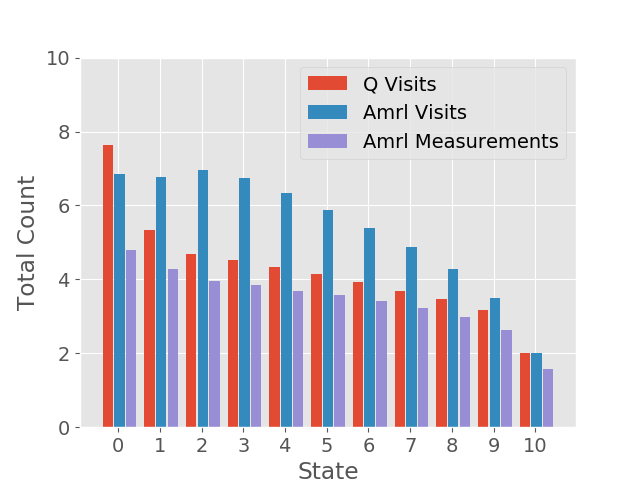}
    \includegraphics[scale=0.25]{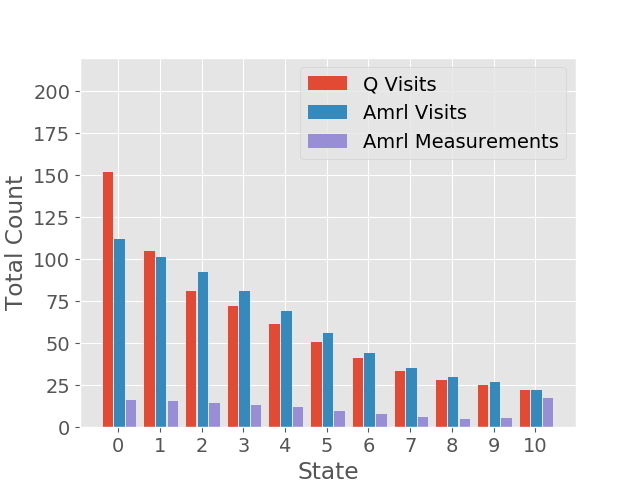}
    \includegraphics[scale=0.25]{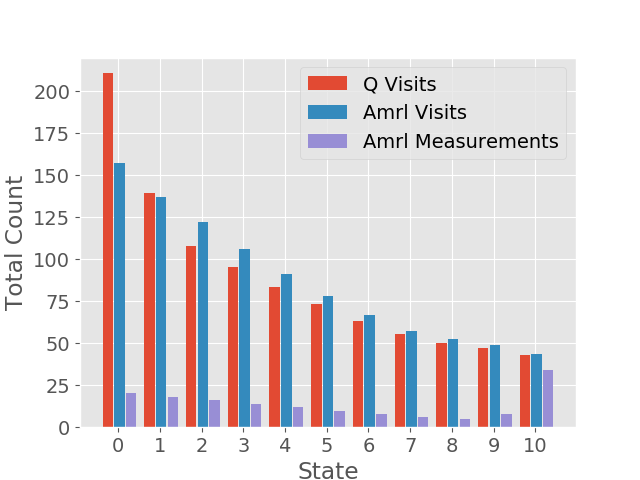}
\caption{Bar plots comparing the state visit and measurement distribution for Q-learning and Amrl-Q on the Chain environment after 1, 20 and 40 (left, centre, right) episodes of training.}
\label{fig:detChainMeasureAnalysisBar}
\end{figure*}

\begin{figure*}
\centering
    \includegraphics[scale=0.4]{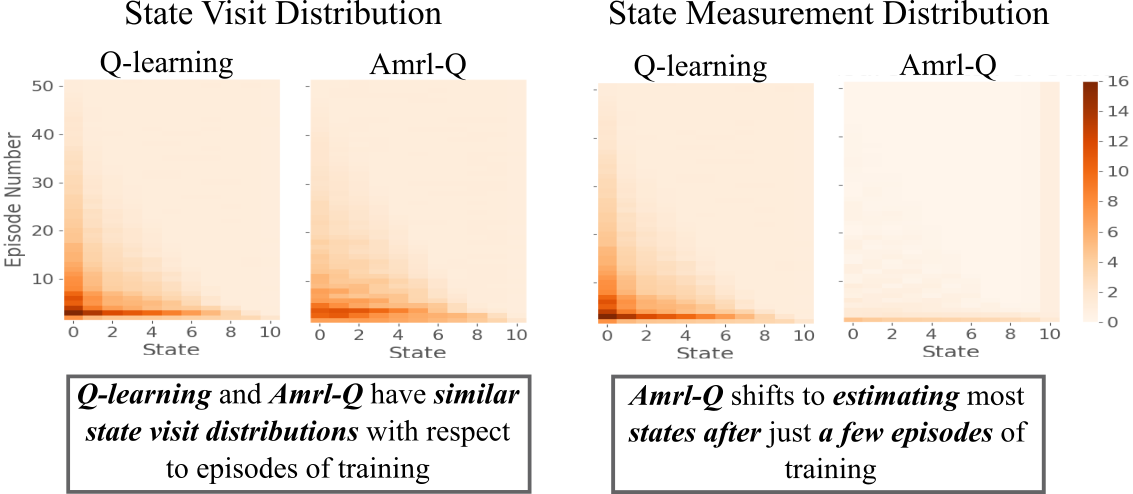}
\caption{Two-dimensional histograms comparing of the number of state visits and measurements made by Q-learning versus Amrl-Q on the Chain environment.}
\label{fig:detChainMeasureAnalysis}
\end{figure*}

Figure \ref{fig:detChainMeasureAnalysisBar} summarizes the total number of visits and measurements made in each state after 1, 20 and 40 episodes of training for Q-learning\footnote{The state visits and measurements are equivalent for Q-learning.} and Amrl-Q. The plot on the left shows that initially Amrl-Q (blue bar) visits most states slightly more frequently than Q-learning (red bar). Importantly, however, the purple bars show that it measures each state less frequently than Q-learning. Thus, the measurement costs are lower from the outset. After 20 and 40 episodes of training (centre and right plots), the state visit frequency of Amrl-Q is consistent with Q-learning. In these later episodes of training, however, Amrl-Q requires significantly fewer state measurements than Q-learning. This highlights the advantage that the Amrl framework has in its ability to shift from measuring the state of the environment to estimating it as more experience (episodes of training) is gathered. This behaviour is shown in greater detail in Figure \ref{fig:detChainMeasureAnalysis}.

Figure \ref{fig:detChainMeasureAnalysis} contains four 2-dimensional histograms. These depict the number of visits to each state (plots 1 and 2) and the number of measurements in each state (plots 3 and 4)\footnote{Q-learning measures the state on each visit, therefore, plots 1 and 3 are the same.} as a function of episodes of training. The $x$-axis specifies the state in the chain and the $y$-axis indicates the number of episodes of training completed. The darker black cells indicate more visits / measurements, whilst white indicates a moderate number and red depicts a low number. As a result of the learning behaviour of Q-learning that was discussed in Section \ref{sec:Amrlq}, the lower diagonal of the state visit and measurement plots for Q-learning, and the state visit plot for Amrl-Q have a light red to black shading, with the darkest black appearing in the lower left corner. The upper diagonal, where the shading is uniformly dark red, shows the time at which the agent has learned a policy that enables it to directly transition from this current state to the goal. This occurs within just a few episodes of training for Q-learning in state 10 (first plot, lower right), whereas it takes approximately 50 episodes of training for state 0 (first plot, upper left). Whilst the state visit distributions are very similar for Q-learning and Amrl-Q, their state measurement distributions have an outstanding difference in magnitude. The max state measurement value for Amrl-Q (right most plot) is 6, in comparison to 16 for for Q-Learning. Moreover, in shading in the Amrl-Q measurement plot quick shift from light red to dark red. In fewer than 30 episodes of training, the agent is able to replace all measurements of the environment with its own estimate. 

\subsection{Frozen Lake $8 \times 8$}
\begin{figure*}
\centering
    \includegraphics[scale=0.4]{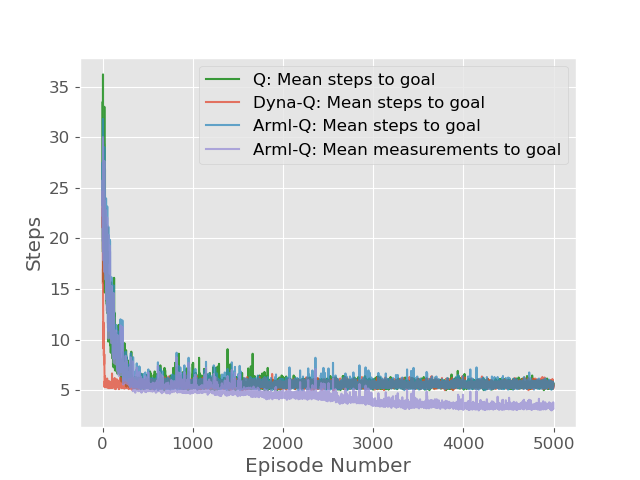}
    \includegraphics[scale=0.4]{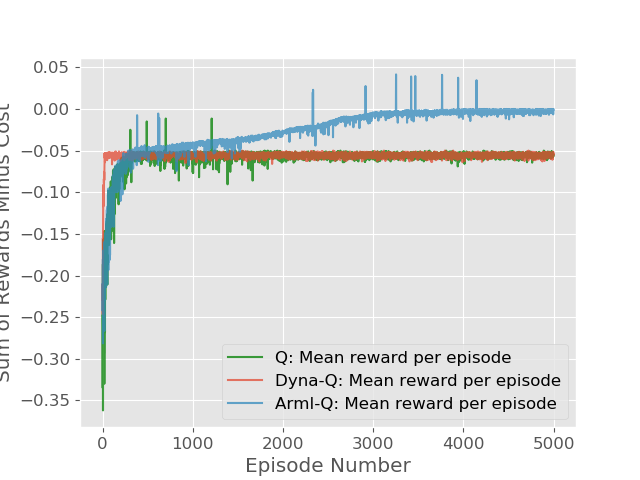}
\caption{Left: Average number of steps to the goal by episode in the deterministic Frozen Lake environment. Right: Average costed return in the deterministic Frozen Lake environment.}
\label{fig:detFrozenLakeStepsAndSum}
\end{figure*}

Figure \ref{fig:detFrozenLakeStepsAndSum} shows the mean number of steps to the goal for each algorithm on the deterministic frozen lake. Similar to the deterministic chain, Amrl-Q learns at the same rate as Q-learning. It takes approximately the same number of steps per episode (green versus blue line). Dyna-Q learns faster than the alternatives, but converges to a similar mean number of steps as Q-learning and Amrl-Q (red line). Amrl-Q requires fewer measurements on average (purple line). The mean number of steps per episode at the end of training for each method is: random agent = 31.95, Q-Learning = 13.99, Dyna-Q = 15.45, Amrl-Q Steps = 18.52. Importantly however, Amrl-Q only takes a mean of 10.50 measurements per episode.

\subsection{Taxi}
\begin{figure*}
\centering
    \includegraphics[scale=0.4]{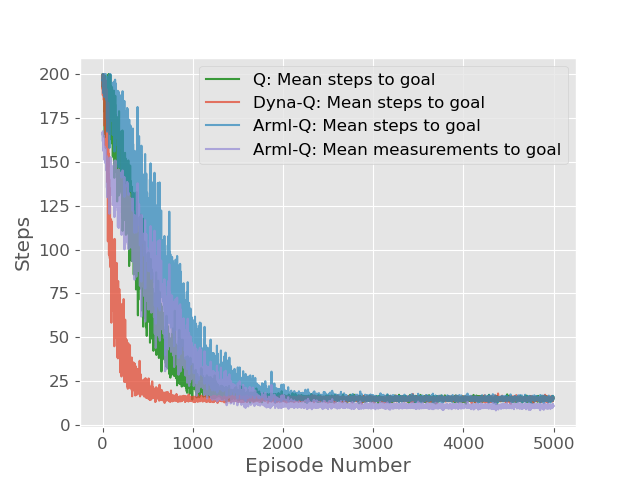}
    \includegraphics[scale=0.4]{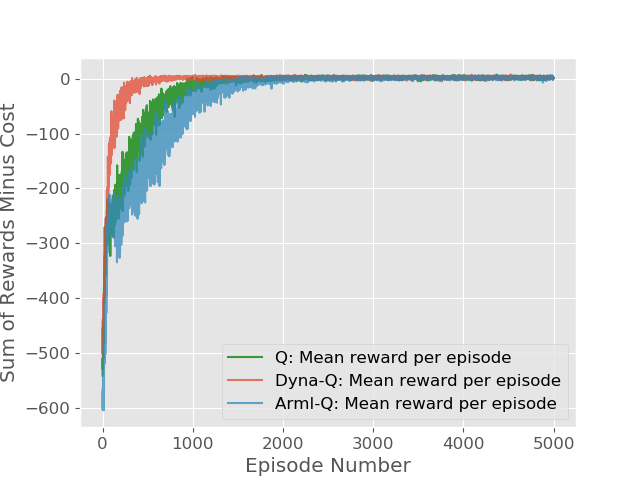}
\caption{Left: Mean number of steps to the goal by episode in the Taxi environment. Right: Mean of the costed return in the Taxi environment.}
\label{fig:detTaxiStepsAndSum}
\end{figure*}

Figure \ref{fig:detTaxiStepsAndSum} depicts the mean number of steps to the goal for each algorithm on the Taxi environment. This is a more challenging environment because it requires the agent to learn an intermediate goal. Nonetheless, the relative performance of the considered algorithms is consistent with our previous results. Amrl-Q learns at a similar rate to Q-learning, and takes approximately the same number of steps (green versus blue line). Dyna-Q learns faster (red line), but converges to a similar average number of steps as Q-learning and Amrl-Q. Amrl-Q requires fewer measurements on average (purple line). The mean number of steps per episode are as follows: random agent = 31.95, Q-Learning = 14.83, Dyna-Q = 14.67, Amrl-Q Steps = 15.30. Amrl-Q take an average of 12.13 measurements per episode.

\subsection{Junior Scientist}

\begin{figure*}
\centering
    \includegraphics[scale=0.4]{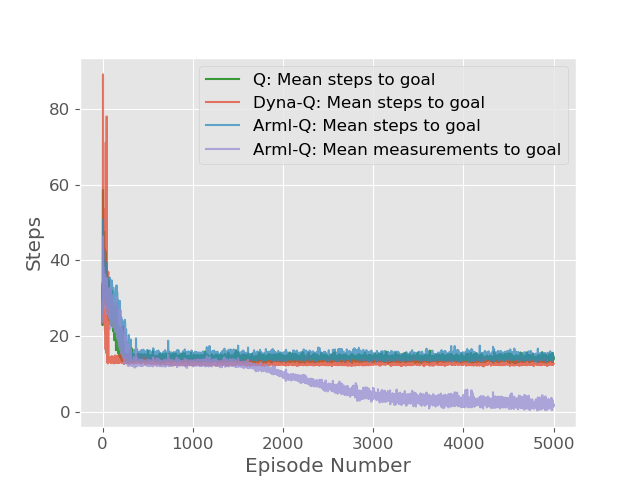}
    \includegraphics[scale=0.4]{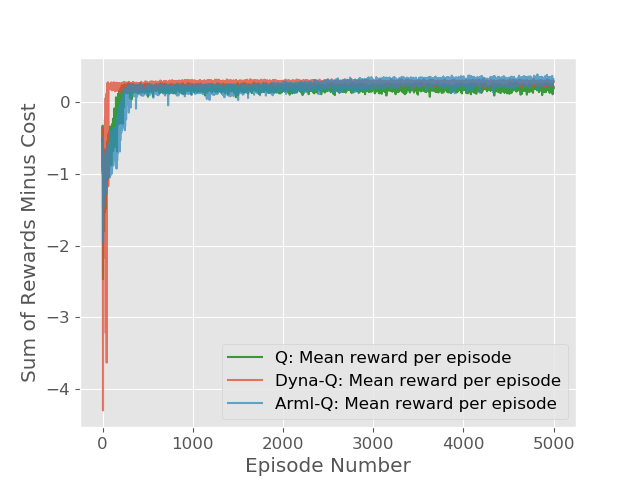}
\caption{Left: Mean steps per episode in the Junior Scientist environment. Right: Mean of the costed returns in the Junior Scientist environment.}
\label{fig:JuniorScientistStepsAndSum}
\end{figure*}

Figure \ref{fig:JuniorScientistStepsAndSum} shows the mean of the costed return for each algorithm on the Junior Scientist environment. Once again, Dyna-Q learns slightly faster than Q-learning and Amrl-Q. The plot on the left clearly shows Amrl-Q shifting away from measuring the state after approximately 2,000 episodes of training (purple line). The fact that the mean steps (blue line) is stable during this shift indicates that the agent is not becoming `lost' in the state space due to bad estimates. 

\section{Discussion}
\begin{figure*}
\centering
    \includegraphics[scale=0.2]{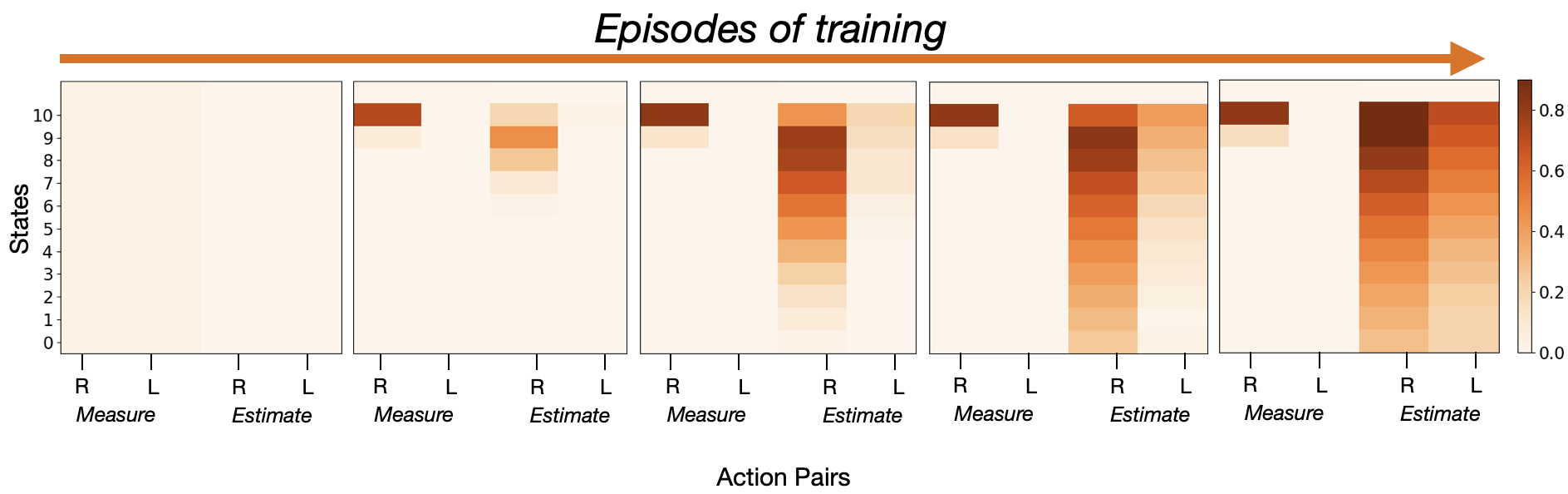}
    \caption{Evolution of the values of the Q-table for Amrl-Q on the deterministic Chain environment.}
\label{fig:qEvolution}
\end{figure*}

Figure \ref{fig:qEvolution} shows the evolution of the values of the Q-table for Amrl-Q over episodes of training on the deterministic chain environment. The $x$-axis shows the four action pairs [(\emph{move left, measure}), (\emph{move right, measure}), (\emph{move left, estimate}), (\emph{move right estimate})] that the agent chooses from. The $y$-axis shows each state, where 0 is the start state and 10 is the goal state. From left to right, the first plot is the initialized Q-table. It is followed by the Q-values after increments of 29 episodes of training. In earlier episodes of training, the action pair (\emph{move right, measure}) has the highest values. The sequence of plots demonstrates that over episodes of training, the action pair (\emph{move right, estimate}) comes to have the highest value. Thus, the agent shifts over time away from its reliance of more costly measurements. 

The shift to estimating the next state occurs naturally within the Q-learning backup algorithm and sufficient exploration. There is a clear trade-off in this evolution. If an agent in state $s$ relies on its state estimator $\hat{P}$ before it is sufficiently accurate, it will be misinformed about its current location. As a result, it is likely to select the wrong action and take more time to reach the goal. Moreover, the agent's Q updates will be applied to the wrong state. Alternatively, if an agent in state $s$ utilizes measurements $m=1$ longer than is necessary (\textit{i.e.}, when $\hat{P}$ is sufficiently accurate), it needlessly pays the measurement cost which lowers its reward. In Amrl-Q, proper exploration and the initialization of the Q-table serve to balance this trade-off. However, more sophisticated solution using model confidence are expected to produce even better performance. We leave the study of such methods to future work. 

\begin{figure*}
\centering
    \includegraphics[scale=0.3]{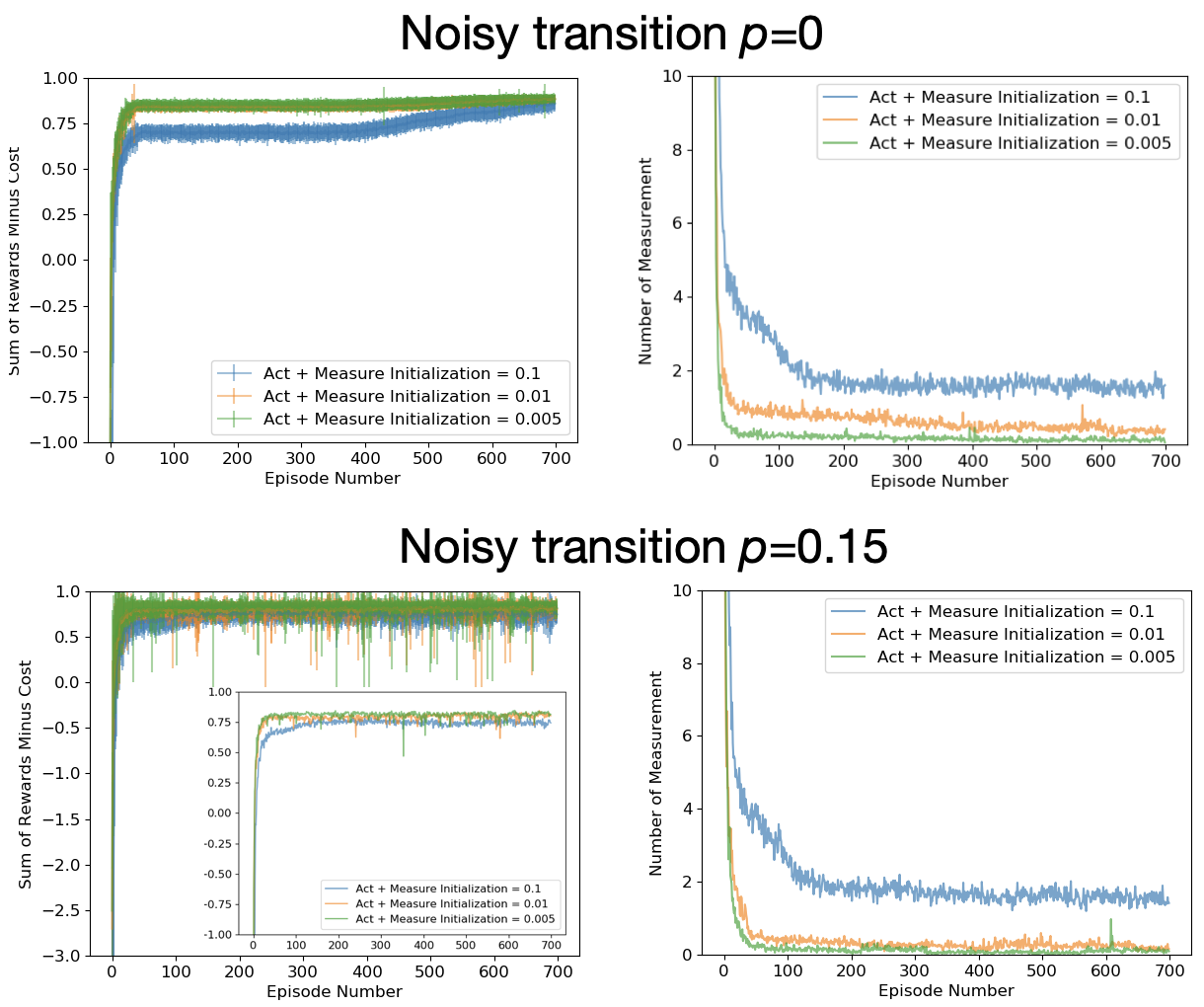}
    \caption{This figure demonstrates how the initialization of the Q-values for \emph{act+measure} affects the number of measurements made by the agent (right column), and how it impacts the costed return in noisy environments.}
\label{fig:measurementsDueToBias}
\end{figure*}

The right column of Figure \ref{fig:measurementsDueToBias} depicts how the initialization of the Q-values associated with measure $m=1$ in Amrl-Q shapes the number of measurements made by the agent. The top plot depicts the number of steps on the deterministic chain and the bottom for the stochastic chain. The episodes of training are plotted on the $x$-axis and the mean number of measurements is plotted on the $y$-axis. This clearly shows that as the initialization is decreased towards zero, the number of measurements made by the agent reduces.  

The number of measurements per state-action pair has important implications on performance in the stochastic environments. In the lower right plot, which applies to the stochastic environment, the difference between the initialization of 0.01 and 0.005 is much smaller than in the deterministic case. In that case, the agent using the initialization of 0.005 shifts to using its state estimator before it is sufficiently accurate. As result, the agent is operating from error prone estimates of is current state, and thus, requires more steps and more measurements on average. 

The column on the left shows how the initialization impacts the costed return. The upper plot shows that given enough time, the agent overcomes the larger initialization to achieve an equivalent costed return as agents with smaller initial values. The lower plot demonstrates the benefit of a large initial value in environments with stochastic transitions. From early episodes of training the difference in mean performance (shown without error bars in the embedded plot) of the agents with different initialization is small. In the large plot (with error bars) it is clear that the larger initial value leads to a notably lower standard deviation. Given the added robustness of the larger initial values, and the fact that the agent will converge to the same performance, we advise against setting it too close to zero.


\begin{figure*}
\centering
    \includegraphics[scale=0.4]{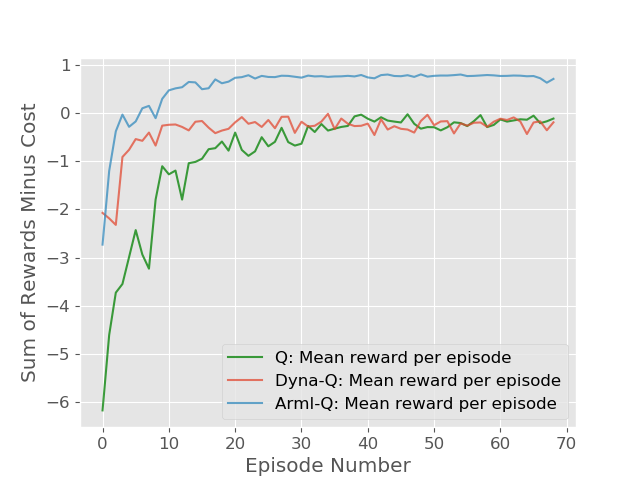}
    \includegraphics[scale=0.4]{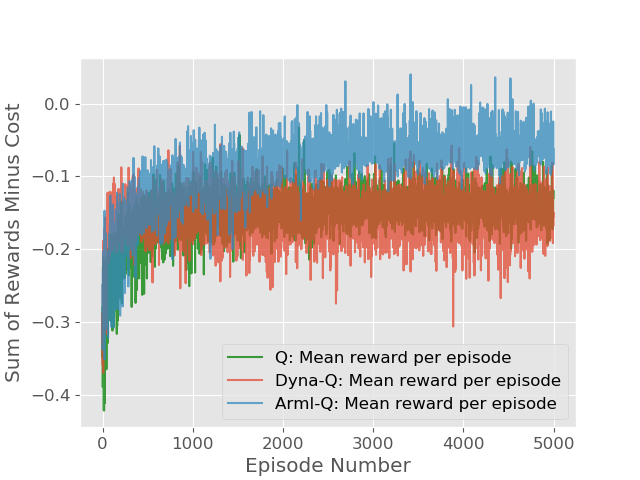}
\caption{Left: Mean of the costed return on the stochastic Chain environment. Right: Mean of the costed return on the slippery Frozen Lake $8 \times 8$ environment.}
\label{fig:stochasatic}
\end{figure*}

The plot on the left in Figure \ref{fig:stochasatic} depicts the mean of the costed return for Amrl-Q, Q-learning and Dyna-Q on the stochastic Chain. In this case, the action pairs involving measure $m=1$ are initialized to 0.01. The results for Slippery Frozen Lake environment are plotted on the right. This is much more complex than the stochastic chain because it involves a larger number of actions and more variability in the transition dynamics. In this setting all methods have a high variance. The actions pairs associated with measure $m=1$ must be set to a large value (in this case 10.0) in order to provide $\hat{P}$ time to stabilize. The Amrl-Q agent begins to slowly shift way from relying on measurements after approximately 1,000 episodes of training.

\section{Conclusion}

We introduced a sequential decision making framework, Amrl, in which the agent selects both an action and an observation class at each time step. The observation classes have associated costs and provide information that depends on time and space. We formulate our solution in terms of active learning, and empirically show that Amrl-Q learns to shift from relying on costly measurements of the environment to using its state estimator via online experience. Amrl-Q learns at a similar rate to Q-learning and Dyna-Q, and achieves a higher costed return. Amrl has the potential to expand the applicability of RL to important applications in operational planning, scientific discovery, and medical treatments. To achieve this, additional research is required to develop Amrl methods for continuous state and action environments, and function approximation methods, such as Gaussian processes and deep learning. 





\bibliographystyle{plain}
\bibliography{library}

\end{document}